\documentclass[10pt,twocolumn,letterpaper]{article}

\usepackage[pagenumbers]{cvpr} 

\usepackage{colortbl}
\usepackage{multirow}
\usepackage{pifont}
\usepackage{makecell}
\usepackage{bm}
\usepackage{graphicx}

\definecolor{cvprblue}{rgb}{0.21,0.49,0.74}
\usepackage[pagebackref,breaklinks,colorlinks,allcolors=cvprblue]{hyperref}

\def\paperID{1} 
\def\confName{CVPR}
\def\confYear{2025}

\definecolor{newgreen}{HTML}{b1dba2}
\definecolor{newblue}{HTML}{a1c1f7}
\definecolor{newpink}{HTML}{FB3199}
\newcommand{\llm}{\textcolor{black}{LLM}}
\newcommand{\lvlm}{\textcolor{black}{LVLM}}
\newcommand{\gt}[1]{\textcolor{gray!75}{#1}}

\title{Zero-shot Action Localization via the Confidence of
Large \\Vision-Language Models}

\author{Josiah Aklilu \\
Stanford University\\
{\tt\small josaklil@stanford.edu}
\and Xiaohan Wang \\
Stanford University\\
{\tt\small xhanwang@stanford.edu}
\and Serena Yeung-Levy \\
Stanford University\\
{\tt\small syyeung@stanford.edu}
}

\begin{document}
\maketitle
\begin{abstract}
Precise action localization in untrimmed video is vital for fields such as professional sports and minimally invasive surgery, where the delineation of particular motions in recordings can dramatically enhance analysis. But in many cases, large scale datasets with video-label pairs for localization are unavailable, limiting the opportunity to fine-tune video-understanding models. Recent developments in large vision-language models (\lvlm) address this need with impressive zero-shot capabilities in a variety of video understanding tasks. However, the adaptation of \lvlm s, with their powerful visual question answering capabilities, to zero-shot localization in long-form video is still relatively unexplored. To this end, we introduce a true \textbf{ZE}ro-shot \textbf{A}ction \textbf{L}ocalization method (\textbf{ZEAL}). Specifically, we leverage the built-in action knowledge of a large language model (\llm) to inflate actions into detailed descriptions of the archetypal start and end of the action. These descriptions serve as queries to \lvlm~ for generating frame-level confidence scores which can be aggregated to produce localization outputs. The simplicity and flexibility of our method lends it amenable to more capable \lvlm s as they are developed, and we demonstrate remarkable results in zero-shot action localization on a challenging benchmark, without any training. Our code is publicly available at \href{https://github.com/josaklil-ai/zeal}{https://github.com/josaklil-ai/zeal}.
\end{abstract}    
\section{Introduction}
\label{sec:intro}

\begin{figure}[h]
    \centering
    \includegraphics[width=\columnwidth, clip, trim=22em 12em 22em 11em]{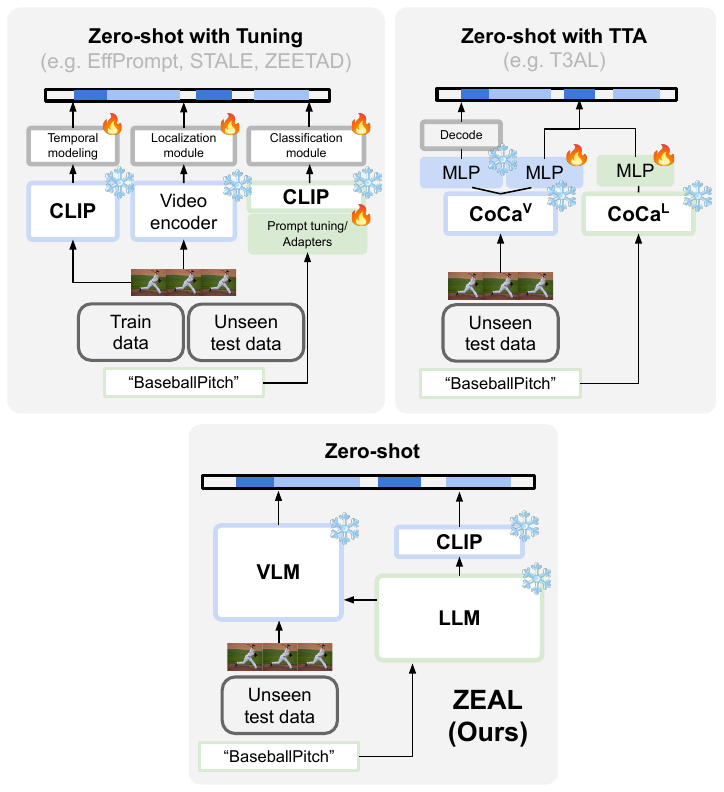}
    \caption{Our framework is a true zero-shot method for action localization in long-form, untrimmed video.
    We leverage off-the-shelf \llm~and \lvlm~for localizing arbitrary actions without needing to optimize any task-specific modules.}
    \label{fig:pull}
\end{figure}

Localization of actions in untrimmed video \cite{unloc, zstad, zeetad, effprompt, afsd, stale, multimodalprompt} in the wild is an active area of interest in a variety of fields. From professional sports analysis to phase identification in minimally invasive surgery, the delineation of the start and end of certain actions in long-form video or temporal action localization (AL), is an integral part of video understanding, enabling precise identification of key moments or adverse events. Unlike action recognition or classification, localization requires a fine-grained understanding of visual inputs in a temporal context, which can be ambiguous if action classes are poorly defined or activities appear relatively similar. Thus, to train high-quality video models, large datasets with extensive ground truth localization annotations are required. Many early AL works \cite{endtoend, actionformer, ssn, rethinkfasterrcnn, asl, tags, tridet, tridetplus} that take a fully-supervised approach to AL have task-specific modules for generating strong action interval candidates. However, fully-supervised approaches are limited in performance on videos out-of-distribution from standard AL benchmarks \cite{thumos14, multithumos, activitynet}. Moreover, by relying on visual features that are not generalizable to unseen video data, these methods break down on in-the-wild videos, thus limiting their application. 

More recently, several lines of work have attempted to address this issue in the zero-shot setting \cite{effprompt, stale, unloc, zeetad}, where not all action categories that are present in testing data are seen during training. The most common approach is localization in zero-shot with tuning, where frozen CLIP \cite{clip} vision and text encoders are combined with learnable prompt vectors \cite{effprompt} and modules \cite{stale, zeetad} to identify action intervals. However, these works are constrained to CLIP pretraining, which provides high-level semantic alignment between whole images and text. Although CLIP exhibits strong performance in cross-modal alignment tasks, AL using CLIP features alone largely ignores developments in more fine-grained alignment for local details. Recent advances in large vision-language models (\lvlm) have demonstrated remarkable potential for dense visual reasoning tasks like grounded visual question answering \cite{llavagrounding, glipv2}, GUI navigation \cite{cogagent} and other cross-modality tasks \cite{cogvlm}. These multi-modal models have strong visual localization capabilities that enable intricate associations between visual elements and textual semantics. 

Although sometimes limited in multi-image or video comprehension, \lvlm's ability to answer basic visual questions compliments the the observation that we as humans naturally follow a similar process when seeking the boundaries of particular actions in videos. When presented with a novel action and tasked to find it in long-form video, there is a tendency to define salient portions of this action via language: \textit{``what does the start of a baseball swing look like?"} $\rightarrow$ \textit{``a bat angled upward, eyes on the ball, knees slightly bent"} or \textit{``how do you know a basketball dunk is complete?"} $\rightarrow$ \textit{``person on two feet, ball on ground, person facing away from hoop"}. Humans then look for these defining moments in video at the frame-level and localize actions. Thus, we posit that action localization can be broken down into two steps: one where actions are deconstructed into their components similar in spirit to \cite{multimodalprompt, zsvmr} (i.e. how the start of the action should look, how the end of the action should look, etc.), and the second where visual-temporal inputs are parsed largely guided via language.  

In this work, we aim to leverage LVLMs to reflect this decomposition and reasoning process. Leveraging the knowledge of everyday actions from a \llm, we decompose generic action classes into precise descriptors of the beginning and the end of these activities. We use these descriptors as queries for an \lvlm~to parse visual information per frame and generate confidence scores. Specifically, we prompt \lvlm~to answer affirmative or negative to questions at the frame-level and extract the model's confidence in answering these visual questions. The resulting scores serve as a proxy for the likelihood of certain orientations and semantics expected at the start of an activity and likewise the end of the activity. From these scores we produce candidate localization outputs by matching start and end times which coincide with the most confident frame-level scores. This framework is agnostic to choice of \llm~for action decomposition in generating descriptors or queries, and to choice of \lvlm~for interpreting visual inputs and producing confidence scores (see \cref{fig:pull}). 

Our paper contributes \textbf{1)} A novel framework leveraging \llm~for action decomposition and \lvlm~for parsing visual inputs to perform true \textbf{ZE}ro-shot \textbf{A}ction \textbf{L}ocalization (\textbf{ZEAL}), \textbf{2)} a new method for creating a surrogate for the confidence of \lvlm~in answering localization queries, and \textbf{3)} without any task-specific training, we achieve competitive or state-of-the-art zero-shot results compared to previous training-based methods on THUMOS14, a representative and challenging AL benchmark of every-day human activities and sports, demonstrating our framework's flexibility in localizing actions in long-form, untrimmed video.

\section{Related Works}
\label{sec:rw}

\subsection{Temporal action localization}
Traditional temporal action localization (AL) delineates the start and end boundaries of actions (and their class) in untrimmed video. There exist many fully-supervised approaches for AL, which can be divided into two-stage approaches \cite{rethinkfasterrcnn, ssn, bsn, bmn, effprompt} and single-stage methods \cite{endtoend, actionformer, tags, temporalmaxer, tridet, tridetplus, asl, adatad}. While two-stage approaches decouple action proposal generation and classification, single-stage approaches do away with noisy proposal detectors and conduct localization in a fully end-to-end manner. \cite{actionformer} presents a strong baseline for AL by using a multi-scale transformer encoder in conjuction with a light-weight convolutional decoder to directly identify action candidates for every frame in the video. \cite{temporalmaxer} uses parameter-free max pooling on strong pretrained features to build a simple backbone for AL. \cite{tridet, tridetplus} models the start and end boundaries of the actions with relative probability distributions. \cite{asl} measures action sensitivity at the class and instance levels and uses a novel contrastive loss to enhance features where positive pairs are action-aware frames and negative pairs are action-irrelevant. \cite{adatad} scales AL with an efficient design allowing for full end-to-end training of a large video encoder. However, all of these methods require extensive supervised training on AL datasets which are limited to the action classes they contain, and thus these methods suffer significant performance drops on unseen action categories in the wild.

\subsection{Zero-shot action localization}
Zero-shot AL works \cite{zstad, tranzad, effprompt, stale, unloc, zeetad} attempt to localize actions of entirely unseen classes at inference time. \cite{effprompt} was the first attempt to leverage CLIP \cite{clip} embeddings to find relevant frames in untrimmed video to the text query for a variety of related localization tasks including moment retrieval and action recognition. The authors demonstrate that the representations from pretrained contrastive models exhibit strong zero-shot generalization and thus can be used for dense video understanding tasks. \cite{stale} also propose a proposal-free method for zero-shot AL by introducing a novel representation masking mechanism for generalization to unseen classes, along with a classification component that runs in parallel to alleviate proposal-error propagation. However, this approach still requires supervised training especially for videos completely outside of the pretraining corpus. Unlike \cite{stale}, our proposed framework leverages the built-in action knolwedge in \llm~ to conduct training free localization. \cite{tranzad} treats zero-shot AL as a direct set-prediction problem, but relies on a 3D convolutional backbone for extracting video features. \cite{unloc} constructs a feature pyramid from a transformer-based video-text fusion module with task specific heads for AL, moment retrieval, and action segmentation. However, this unified architecture still relies solely on CLIP as the primary visual feature extractor and could potentially have poor generalization performance on out-of-domain video datasets. \cite{zeetad} is similar to \cite{stale} in that representation masking is used to localize actions, but a 3D video encoder is used in addition to CLIP visual features to prompt action proposals. 

Moreover, a key preliminary work to ours \cite{t3al} leverages information at test-time to generate and refine localization outputs using pretrained \lvlm. However, \cite{t3al} relies on having many positive and negative samples of frames relevant to action classes so that a self-supervised objective can be learned on unlabelled test data. The inherent tie to sample size at inference limits utility in settings where enough test data is available for adaptation. Our proposed framework enjoys a truly training-free setup that harnesses the capabilities of existing pretrained models for localization.

\subsection{LLMs and LVLMs for video understanding}
Our work is situated amongst works leveraging \llm s and \lvlm s for diverse video understanding tasks \cite{multimodalprompt, simplellm, lita}. For localization in particular, some prior work \cite{multimodalprompt} use \llm~to decompose actions into defining attributes that can be more easily aligned with visual inputs. \cite{simplellm} leverages image captioners at the frame-level and \llm~to conduct long-range video question answering. \cite{lita} introduces slow and fast visual tokens so that \llm~can reason over temporal information. In this work, we do not rely on \llm~for localization, but rather measure the confidence of \lvlm~in understanding visual inputs for fine-grained activity detection.

Our method uses an entirely different approach to AL leveraging \llm~as to decompose actions into language queries for \lvlm~to parse visual inputs, augmenting action class labels while retaining the robustness of pretrained vision encoders, enabling strong generalization capabilities on a variety of video datasets.
\section{Method}
\label{sec:method}

In the following sections, we introduce \textbf{ZEAL}, a generalist framework for AL in untrimmed video using \llm~ and \lvlm~ for action decomposition and visual perception, respectively. Our approach is flexible, allowing for each component to be exchanged with more capable models as they are made available to the community.

We consider the following action localization setting: given a video $v_i \in \mathbb{R}^{T \times 3 \times H \times W}$, $1 \leq i \leq N$ and where $T$ is the number of frames for the video and is variable, we wish to localize actions $a_j = (t_j^s, t_j^e, y_j)$, where $t_j^s, t_j^e$ are the start and end times ($1 \leq t_j^s < t_j^e \leq T$), $j$ is the number of action instances in video $v_i$ ($1 \leq j \leq M$), and each action comes from a closed-set vocabulary (i.e. $y_j \in \{1, \dots, K\}$ and $K$ is the number of unique action classes). $M$ can be large, and there can be multiple action classes represented in a single video. 

\subsection{Generating Vision-Language Queries for Action Localization}
\label{stage1}
The first stage of our framework involves generating queries for the \lvlm~to appropriately identify key markers of the actions in question for localization. To this end, we employ an \llm~(e.g. GPT-4o \cite{gpt4o}) to decompose action classes into two questions, one seeking what the archetypal \textit{start} of the action should look like ($q^s$), and the other what the archetypal \textit{end} of the action should look like ($q^e$). Importantly, we enforce that these questions are directly ``yes/no" answerable and prompt the \lvlm~in subsequent stages to only answer affirmative or negative. Prior work \cite{visualdescriptors} has successfully demonstrated that the decomposition of class categories into semantic descriptors via mining of an \llm~enhances image classification. We employ a similar idea here by extracting the built-in action knowledge of \llm~to inflate action classes into descriptors that delineate the boundaries of these actions. This allows the \lvlm~to parse its visual inputs and subsequently localize actions. Moreover, we prompt the \llm~to give a short description of the action ($q^a$) that will be used to construct a CLIP-based actionness score (see Section \ref{stage2}). Here is an example of a produced start and end query, along with a short description, for the \textit{``CliffDiving"} action:

\begin{quote}
    \ttfamily
    $q^s$: Is the person standing on the edge of a cliff above water? \\
    $q^e$: Is the person submerged in or breaking the surface of the water? \\
    $q^a$: A person jumps from a cliff into water.
\end{quote}

If all action classes are known \textit{a priori}, these queries need only be generated once, incurring minimal costs if using proprietary APIs even with large $K$.

\subsection{Coarse Action Class Filtering for Relevancy}
\label{stage2}
To reduce computational overhead introduced by repeated forward passes through the \lvlm, we implement an instance-level filtering of action classes that are relevant to the untrimmed video. This is done primarily to reduce the search space for \lvlm~question answering. The instance-level filtering is implemented as follows: given a closed set of $K$ action classes, we sample frames from the video $v_i$ and use CLIP \cite{siglip} to embed these frames. We collect the top-\textit{k} action classes (cosine similarity w.r.t image features) most likely to occur in a video containing these frames (see Section \ref{filtering} for more details).

We then only consider these $\hat{K} << K$ action classes for the rest of the pipeline. For example, some datasets may have hundreds of action classes, but only a few action instances per video, so this coarse-filtering of action classes dramatically decreases the number of queries needed to conduct localization. 

\subsection{CLIP-based Actionness}
\label{stage3}
We also utilize the short description $q^a$ produced by the \llm~for a given action and measure the cosine similarity ($\rho_t$) to frame-level CLIP \cite{clip} features. Again, we use a CLIP vision tower for extracting frame-level features and the corresponding text tower to embed the description. The cosine similarities are used when generating action interval proposals (see Section \ref{stage5}) to avoid creating intervals containing background frames. 

\subsection{Fine-grained Localization with LVLM Outputs}
\label{stage4}
After filtering actions to $\hat{K}$ classes for a given video $v_i$, we then proceed to query the \lvlm~and produce confidence scores per frame per query per filtered class ($1 < k < \hat{K}$). Particularly, for each frame of video, we prompt \lvlm~with the image and $q_k^s$ or $q_k^e$ and observe the language response. For example, an input to the \lvlm~could be:

\begin{quote}
    \ttfamily
    <IMG> This is a frame from a video of \{$k$\}.\{$q_k^s$\} Only answer yes or no.
\end{quote}

\begin{figure*}[h]
    \centering
    \includegraphics[width=\textwidth, trim=6em 1em 6em 2em]{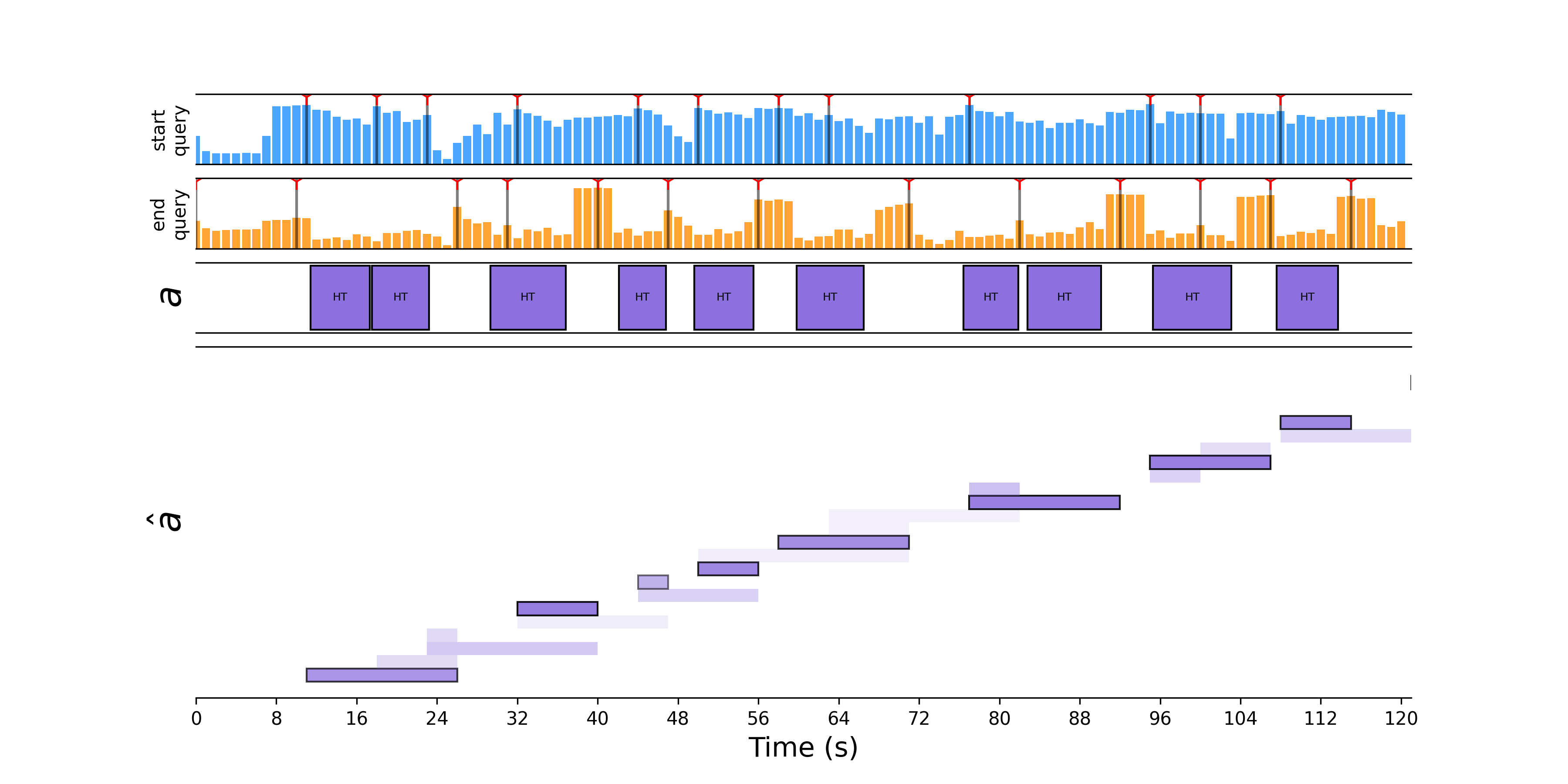}
    \caption{Visualization of soft-scores associated with $q^s$ (blue) and $q^e$ (orange) for a video in the THUMOS14 test set, where the ground truth action are \textit{``Hammer Throw"}. In some cases, there is a striking distribution peak where the LVLM is confidently affirmative in answering $q^s$ (such as around second 96) and $q^e$ (seconds 104-108). At 1 fps, the distribution of these scores has a rough alignment to the beginning and end of the ground truth actions. The light gray lines overlaying the barplots indicate the chosen start and end candidates (based on top-$p\%$ threshold). The predictions $\hat{a}$ are constructed from matching start and end candidates as described in Section \ref{stage5}. Highly confident predicted intervals are outlined for illustration, and we show the first 2 minutes of the video for clarity.}
    \label{fig:qual}
\end{figure*}

At the first step of generation, we extract the logits of the "yes" token ($l_y$) and the "no" token ($l_n$) and compute two scores (one for $q^s$ and $q^e$) for frame $t$ as $s_t = e^{l_y} / (e^{l_y} + e^{l_n}) \in [0,1] $. These scores $S_k \in \mathbb{R}^{T \times 2}$ serve as confidence scores of the \lvlm~in affirming the question about the current state of the activity (as defined by $q_k^s$ or $q_k^e$). This also results in two disparate distributions, one which peaks at the beginning of action instances in the untrimmed video, and one that peaks at the ends of these instances (see \cref{fig:method} and \cref{fig:qual}).

\subsection{Creating Action Interval Proposals}
\label{stage5}
These confidence scores $S_k$ must be aggregated to produce localization outputs. To integrate information within a temporal context into these confidence scores, we employ a modified version of min-max normalization for each column of scores $s_k \in \mathbb{R}^T$ associated with $q^s$ or $q^e$: \begin{equation}
s_k = \frac{s_k - \text{min}(s_k)}{\text{max}(s_k) - \text{min}(s_k)} (1 - 2 \epsilon) + \epsilon
\end{equation}
where $\epsilon$ controls the resulting range of the scores. This is done to ensure that min-max normalization doesn't result in scores at the extremes 0 or 1. We use $\epsilon = 0.05$ in our experiments.

We then construct candidate action intervals by first identifying the frames within the top-$p\%$ of scores for $q_k^s$ and consider these as action start candidates $\hat{t}_{j}^s$ (and the same for producing end candidates $\hat{t}_{j}^e$ by $q_k^e$ soft-scores). Each start candidate is paired with each end candidate to produce $M^* > M$ potentially overlapping intervals $\hat{a}_{j} = (\hat{t}_{j}^s, \hat{t}_{j}^e, y_j)$ throughout $v_i$ (we discard intervals that contain multiple end candidates that occur before another start candidate within the interval). However, this results in multiple overlapping proposals that must be filtered to produce final localization outputs. We assign each proposed interval a score defined as follows:
\begin{equation}
\phi_j = \lambda \phi_j^{\text{conf}} + (1 - \lambda) \phi_j^{\text{actionness}}
\end{equation}
where $\phi_j^{\text{conf}} = s_{\hat{t}^s} + s_{\hat{t}^e}$ approximates the confidence of the localization,  $\phi_j^{\text{actionness}} = \big(\sum_{t = \hat{t}^s}^{\hat{t}^e} \rho_t \big) / \big(\hat{t}^e - \hat{t}^s\big)$ measures the relevancy of frames in the predicted interval to the action of interest (normalized by interval length), and $\lambda$ is a tradeoff parameter. We then use 1-dimensional soft non-maximum suppression \cite{softnms} (with Gaussian penalty controlled by hyperparameter $\sigma$) with scores $\phi$ to discard extraneous proposals. We also merge intervals with high overlap, and the remaining action intervals are the predictions of actions $a_j$ in video $v_i$.

\begin{figure*}
    \centering
    \includegraphics[width=\textwidth, trim=2em 7em 2em 2em]{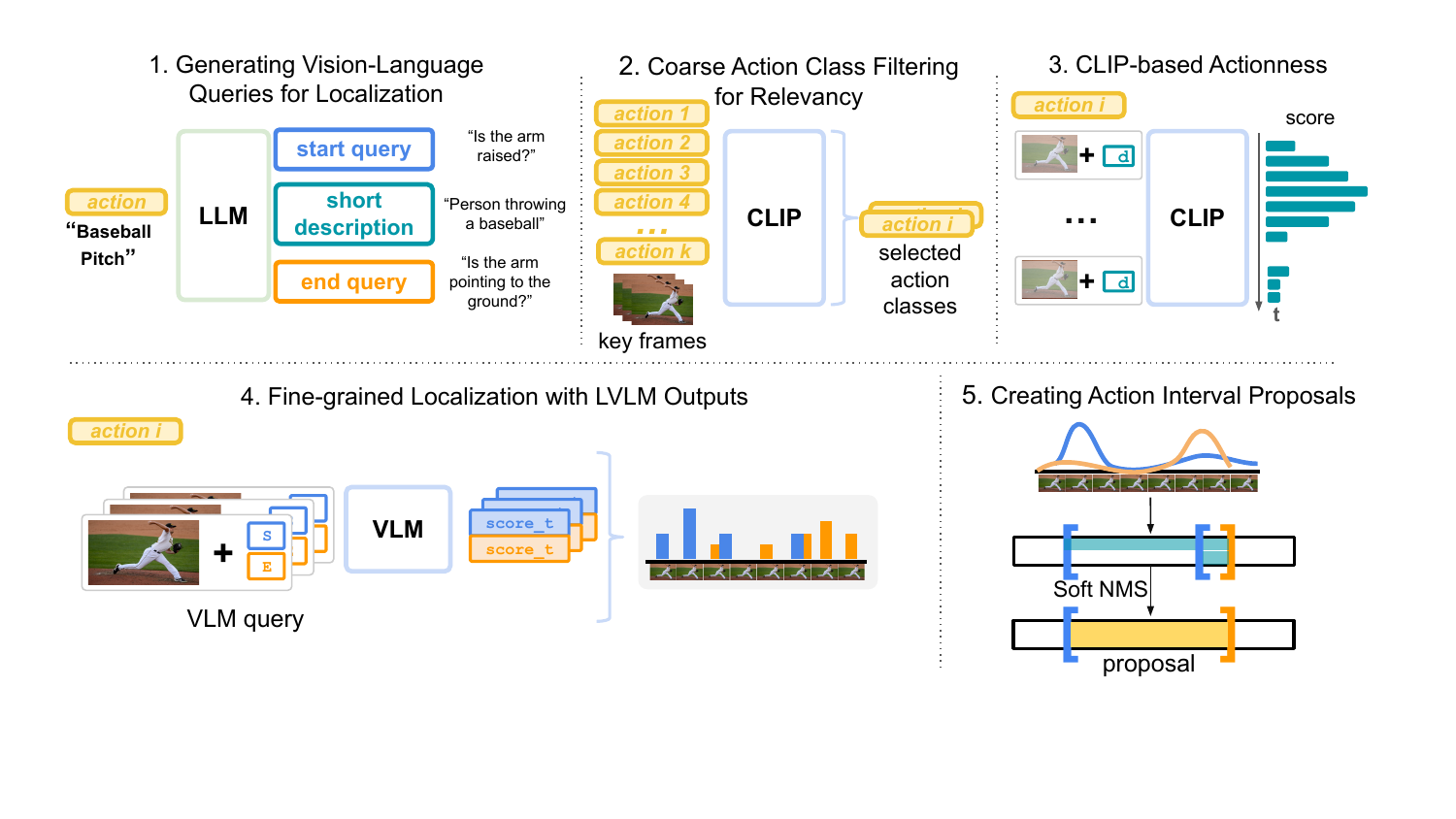}
    \caption{A depiction of the different stages of our framework. \textbf{Stage 1)} Action class names are provided to the \llm~and decomposed into three queries: a start query, and end query, and a short description. These queries are yes/no answerable questions that the \lvlm~uses to assess the likelihood that an activity is starting or ending. \textbf{Stage 2)} To reduce the query search space, we can use a \lvlm~or CLIP to rank all action classes given key frames from a given video to determine which action classes are most likely to be depicted. Only these selected action classes are considered in the rest of the pipeline. \textbf{Stage 3)} We extract frame-wise CLIP similarity scores with the derived action descriptions to supplement interval filtering when proposals are generated. \textbf{Stage 4)} The queries generated by \llm~are then used to prompt the \lvlm~with each frame of video, and confidence scores are extracted at the frame level. \textbf{Stage 5)} Finally, the distribution of confidence scores are used to generated action interval proposals.}
    \label{fig:method}
\end{figure*}

Our framework (see \cref{fig:method}) can leverage \llm~and \lvlm~for action decomposition and visual perception with the ultimate goal of producing action localization outputs. With minimal hyperparameters ($\lambda$ and $\sigma$), our proposed method does not require any parameter optimization or training.

\section{Results}
\label{sec:results}

In the following sections, we demonstrate the efficacy of \textbf{ZEAL} on a standard localization benchmark and investigate different components of the proposed framework, including its influence on interval proposal generation.

\subsection{Datasets and Evaluation Settings}
Following prior art in zero-shot AL, we demonstrate the effectiveness of our method on the THUMOS14 \cite{thumos14} video dataset. THUMOS14 consists of 200 training and 213 test videos of 20 action classes of sports-related activities. We choose this dataset since the videos in this benchmark average a length of 4.4 minutes, much longer than standard AL datasets and more representative of in-the-wild video.

We report the experimental results of previous methods and our proposed framework across two different evaluation settings:
\textbf{(Zero-shot 75\% | 25\%)} where 75\% of action classes with labels appear during training, while the remaining 25\% are only seen at inference and \textbf{(Zero-shot 50\% | 50\%)} where 50\% of action classes appearing during training, and the remaining 50\% at inference.

Performance in this work is measured by the standard mean average precision (mAP) at different intersection-over-union (IoU) thresholds as reported in prior art. For producing splits, we randomly sample 10 subsets of the action classes and report the average over these splits to guarantee statistical significance. Note that our method doesn't use any of the classes in the train splits.

\textbf{Implementation details.} In our work, we use GPT-4o \cite{gpt4o} (\textit{gpt-4o-2024-11-20}) as the \llm~to produce vision-language queries for localization, and we employ two representative \lvlm s that represent the dominant training designs of modern \lvlm: we use Pixtral (12B) \cite{pixtral} as a highly-capable \lvlm~for its ability to reason over fine-grained details in images (due to a new vision encoder trained from scratch to natively handle variable image sizes and aspect ratios), and we also use LLaVA-OneVision (7B) \cite{llavanext, llavaov} representing a class of \lvlm s built on the LLaVA \cite{llava} framework (fine-tuning a visual projector to map visual inputs into the input space of \llm). In our empirical experiments, we find that $\lambda = 0.1$ and $\sigma = 1.3$ works best for the hyperparameters. We conduct all experiments on NVIDIA L40S or NVIDIA RTX A6000 GPUs, processing videos in parallel if multiple GPUs are available.

\begin{table*}[t]
    \centering
    \caption{Zero-shot action localization results on the THUMOS14 benchmark. Results for the two most common zero-shot settings with seen and unseen splits are taken from the respective prior art. EffPrompt, STALE, MMPrompt, and ZEETAD all require training data (\dag~ indicates results taken from \cite{t3al} where these methods are trained with classes out-of-distribution from the test split). T3AL is a test-time adaptation method closest to our work in that it does not use data from the train split, but uses test labels for optimization. Our zero-shot method requires no adaptation or parameter updates of any kind. We also include a fully-supervised baseline for context. The relevant metric is mean average precision (higher is better) under various IoU thresholds. OOD = Out-of-Domain (train and test labels come from different domains), TF = Training Free, NTTO = No Test-time Optimization.}
    \begin{tabular}{c l c c c c c c c c c}
        \toprule 
        \rowcolor{gray!10} & & & & & \multicolumn{6}{c}{THUMOS14}\\
        \midrule
        Setting & Method & OOD & TF & NTTO  & mAP@0.3 & 0.4 & 0.5 & 0.6 & 0.7 & Avg \\
        \midrule
        \midrule 
        
        Fully supervised & \gt{ActionFormer} \cite{actionformer} & \gt{\ding{55}} & \gt{\ding{55}} & \gt{\ding{51}} & \gt{82.1} & \gt{77.8} & \gt{71.0} & \gt{59.4} & \gt{43.9} & \gt{66.8} \\
        
        \midrule
        
        \multirow{9}{*}{\makecell[c]{Zero-shot\\ 75\% Seen, 25\% Unseen}} & \gt{EffPrompt} \cite{effprompt} & \gt{\ding{55}} & \gt{\ding{55}} & \gt{\ding{51}} & \gt{39.7} & \gt{31.6} & \gt{23.0} & \gt{14.9} & \gt{7.5} & \gt{23.3} \\
        
        & \gt{STALE} \cite{stale} & \gt{\ding{55}} & \gt{\ding{55}} & \gt{\ding{51}} & \gt{40.5} & \gt{32.3} & \gt{23.5} & \gt{15.3} & \gt{7.6} & \gt{23.8} \\
        
        & \gt{MMPrompt} \cite{multimodalprompt} & \gt{\ding{55}} & \gt{\ding{55}} & \gt{\ding{51}} & \gt{46.3} & \gt{39.0} & \gt{29.5} & \gt{18.3} & \gt{8.7} & \gt{28.4} \\
        
        & \gt{ZEETAD} \cite{zeetad} & \gt{\ding{55}} & \gt{\ding{55}} & \gt{\ding{51}} & \gt{61.4} & \gt{53.9} & \gt{44.7} & \gt{34.5} & \gt{20.5} & \gt{43.2} \\

        \cmidrule(r){2-11}
        
        & EffPrompt\textsuperscript{\dag} & \ding{51} & \ding{55} & \ding{51} & 7.1 & 5.9 & 4.5 & 3.4 & 2.2 & 4.6 \\
        
        & STALE\textsuperscript{\dag} & \ding{51} & \ding{55} & \ding{51} & 0.5 & 0.3 & 0.2 & 0.2 & 0.2 & 0.3 \\

        & T3AL\textsuperscript{0} \cite{t3al} & \ding{51} & \ding{51} & \ding{51} & 11.1 & 6.5 & 3.2 & 1.5 & 0.6 & 4.6 \\
        & T3AL & \ding{51} & \ding{51} & \ding{55} & 19.2 & 12.7 & 7.4 & 4.4 & 2.2 & 9.2 \\

        \cmidrule(r){2-11}

        \rowcolor{newblue!15} & ZEAL-Pixtral & \ding{51} & \ding{51} & \ding{51} & 18.2 & 11.9 & 7.0 & 3.9 & 1.4 & 8.5 \\
        \rowcolor{newblue!15} & ZEAL-Pixtral\textsuperscript{\ding{59}} & \ding{51} & \ding{51} & \ding{51} & 27.4 & 17.7 & 10.5 & 5.6 & 1.9 & 12.6 \\

        \rowcolor{newgreen!15} & ZEAL-LLaVA-OV & \ding{51} & \ding{51} & \ding{51} & 22.1 & 16.1 & 11.0 & 5.7 & 3.0 & 11.6 \\
        \rowcolor{newgreen!15} & ZEAL-LLaVA-OV\textsuperscript{\ding{59}} & \ding{51} & \ding{51} & \ding{51} & \textbf{35.1} & \textbf{23.8} & \textbf{15.2} & \textbf{7.6} & \textbf{3.5} & \textbf{17.0} \\
        
        \midrule
        
        \multirow{9}{*}{\makecell[c]{Zero-shot\\ 50\% Seen, 50\% Unseen}} & \gt{EffPrompt} \cite{effprompt} & \gt{\ding{55}} & \gt{\ding{55}} & \gt{\ding{51}} & \gt{37.2} & \gt{29.6} & \gt{21.6} & \gt{14.0} & \gt{7.2} & \gt{21.9} \\
        
        & \gt{STALE} \cite{stale} & \gt{\ding{55}} & \gt{\ding{55}} & \gt{\ding{51}} & \gt{38.3} & \gt{30.7} & \gt{21.2} & \gt{13.8} & \gt{7.0} & \gt{22.2} \\
        
        & \gt{MMPrompt} \cite{multimodalprompt} & \gt{\ding{55}} & \gt{\ding{55}} & \gt{\ding{51}} & \gt{42.3} & \gt{34.7} & \gt{25.8} & \gt{16.2} & \gt{7.5} & \gt{25.3} \\
        
        & \gt{ZEETAD} \cite{zeetad} & \gt{\ding{55}} & \gt{\ding{55}} & \gt{\ding{51}} & \gt{45.2} & \gt{38.8} & \gt{30.8} & \gt{22.5} & \gt{13.7} & \gt{30.2} \\
        
        \cmidrule(r){2-11}
        
        & EffPrompt\textsuperscript{\dag} & \ding{51} & \ding{55} & \ding{51} & 5.4 & 4.4 & 3.5 & 2.7 & 1.9 & 3.6 \\
        
        & STALE\textsuperscript{\dag} & \ding{51} & \ding{55} & \ding{51} & 1.3 & 0.7 & 0.6 & 0.6 & 0.4 & 0.7 \\

        & T3AL\textsuperscript{0} \cite{t3al} & \ding{51} & \ding{51} & \ding{51} & 11.4 & 6.8 & 3.5 & 1.7 & 0.6 & 4.8 \\
        & T3AL & \ding{51} & \ding{51} & \ding{55} & 20.7 & 14.3 & 8.9 & 5.3 & 2.7 & 10.4 \\
        
        \cmidrule(r){2-11}
        
        \rowcolor{newblue!15} & ZEAL-Pixtral & \ding{51} & \ding{51} & \ding{51} & 17.0 & 10.9 & 6.3 & 3.5 & 1.2 & 7.8 \\
        \rowcolor{newblue!15} & ZEAL-Pixtral\textsuperscript{\ding{59}} & \ding{51} & \ding{51} & \ding{51} & 26.6 & 16.7 & 9.7 & 5.1 & 1.7 & 11.9 \\

        \rowcolor{newgreen!15} & ZEAL-LLaVA-OV & \ding{51} & \ding{51} & \ding{51} & 21.1 & 15.0 & 9.9 & 5.0 & 2.6 & 10.7 \\
        \rowcolor{newgreen!15} & ZEAL-LLaVA-OV\textsuperscript{\ding{59}} & \ding{51} & \ding{51} & \ding{51} & \textbf{33.8} & \textbf{22.4} & \textbf{14.0} & \textbf{6.8} & \textbf{3.0} & \textbf{16.0} \\
        
        \bottomrule
    \end{tabular}
    \small{\textsuperscript{\ding{59}} indicates using video-level labels (action classes for a video are known) for filtering action classes.}
\label{tbl:thumos}
\end{table*}

\subsection{Filtering Action Classes for Efficiency}
\label{filtering}
To reduce the number of \lvlm~ queries needed to localize actions, we filter relevant action classes to the video of interest using CLIP-ranking of action classes. Specifically, we use EVA-02-CLIP-L/14+ \cite{evaclip} to encode a specific number of frames uniformly sampled throughout video $v_i$ and measure the cosine similarity between the mean image features and embedded action class names. 

\begin{figure}[b]
    \centering
    \includegraphics[width=0.85\columnwidth, trim=0 0 0 2em]{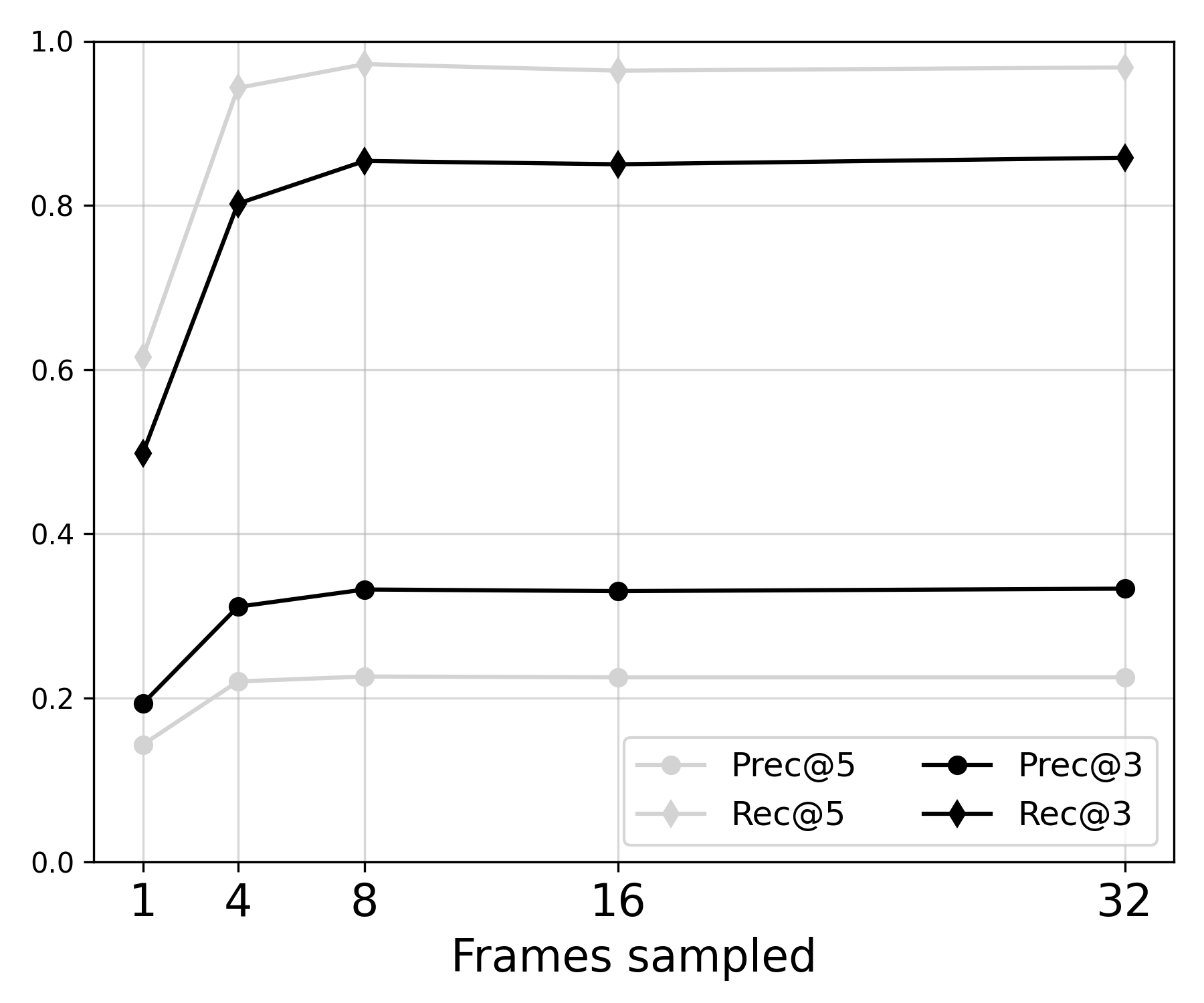}
    \caption{Precision and recall for action class filtering as a function of number of uniformly sampled frames from a particular video. We plot these for $k=5$ (gray) and $k=3$ (black) and note that even with selecting the 3 most likely classes renders adequate recall for generating \lvlm~queries.}
    \label{fig:precrec}
\end{figure}

\vspace{-0.2em}
We record the top-$k$ action classes and only use the $q^s$ and $q^e$ relevant to these action classes when querying \lvlm. For THUMOS14, we find that $k = 3$ is sufficient for high recall in assessing the correct action classes that could be present in video. \Cref{fig:precrec} depicts precision curves and recall curves for different number of frames sampled per video. From this, we also find 8 frames suffices for accurately representing video for this step of our framework.

\subsection{Zero-shot Action Localization}
We compare our method to several baselines in the literature with varying levels of supervision for AL. Prior work in zero-shot AL can be divided into tuning-based zero-shot methods, where some action classes are seen at training time, and test-time adaptation methods, which do not have access to training classes but can leverage test classes. Our method is entirely training-free and does not require localization labels. This reflects real-world scenarios where AL outputs are limited and entirely new action classes are observed in-the-wild.

\Cref{tbl:thumos} benchmarks ZEAL on the THUMOS14 dataset. The grayed-out methods EffPrompt, STALE, MMPrompt, and ZEETAD, are recent zero-shot AL methods. However, these works are trained with AL labels of either 75\% or 50\% of THUMOS14 action classes. The claim that these methods are zero-shot is unsubstantiated when evaluating these methods' performance on AL labels outside of THUMOS14 classes. For example, as observed by \cite{t3al}, EffPrompt has an average mAP of 23.3 on the THUMOS14 test set when trained with 75\% of the action classes, but this drops to 4.6 if the method is trained on HMDB51 \cite{t3al, hmdb51} (a similar performance drop occurs with STALE). Thus, the performance of these methods does not truly capture zero-shot generalization capability. The closest work related to ours (T3AL) attempts to address the out-of-domain problem in prior zero-shot AL work via test-time optimization, but the reliance on retrieving good positive and negative pairs for learning on unlabelled data suffers from false positives. As noted by the authors, T3AL relies on finding informative negative frames (frames containing action-related content but not part of the true action interval) for a single test sample in order to properly conduct adaptation, which suffers from noisy, in-the-wild data where video can contain unrelated video frames. T3AL also includes experiments with zero adaptation steps (T3AL$^0$), however the performance of this method (which is the same zero-shot setup as our work) suffers. 

Our method outperforms T3AL's best baseline up to a 26.1\% improvement in average mAP on the 75\% | 25\% split, and up to a 2.9\% improvement on the 50\% | 50\% split, using only pretrained \lvlm~to affirm visual questions and generate soft-scores, and without parameter updates of any kind. If video-level labels are available (if the set of action classes for a given video is constrained), our method enjoys strong performance gains. We also note that using LLaVA-OneVision as the \lvlm~in our setup performs marginally better than using Pixtral as \lvlm.

\subsection{Ablations}
In addition, we conduct ablations to investigate the contributions of different components of our framework. All ablations reported here are performed on the THUMOS14 benchmark (50\% | 50\% split) and assuming action classes for a given video are known.

\begin{table}
\centering 
\caption{Ablation of backbone for CLIP-based actionness on final localization performance (on zero-shot 75-25 split, parameters held equal).} 
\begin{tabular}{c c}
\toprule 
\rowcolor{gray!10} CLIP & Average mAP \\
\midrule
SigLIP \cite{siglip} & 15.9 \\
\rowcolor{newblue!25} EVA-CLIP \cite{evaclip} & 17.0 \\
\bottomrule
\end{tabular}  
\label{tbl:clip}
\end{table}

\textbf{CLIP for actionness scores.}~ First, we ablate the backbone used for  deriving actionness scores for intervals (as in Section \ref{stage3}). We test EVA-02-CLIP-L/14+ \cite{evaclip} and SigLIP-SoViT-400M/14 \cite{siglip} as strong CLIP models for image and text embedding. We show downstream localization performance in \cref{tbl:clip} when using either backbone for actionness, and we find that EVA-CLIP is the best performing CLIP backbone for our framework.

\begin{figure}[b]
    \centering
    \includegraphics[width=0.95\columnwidth, trim=0 0 0 4em]{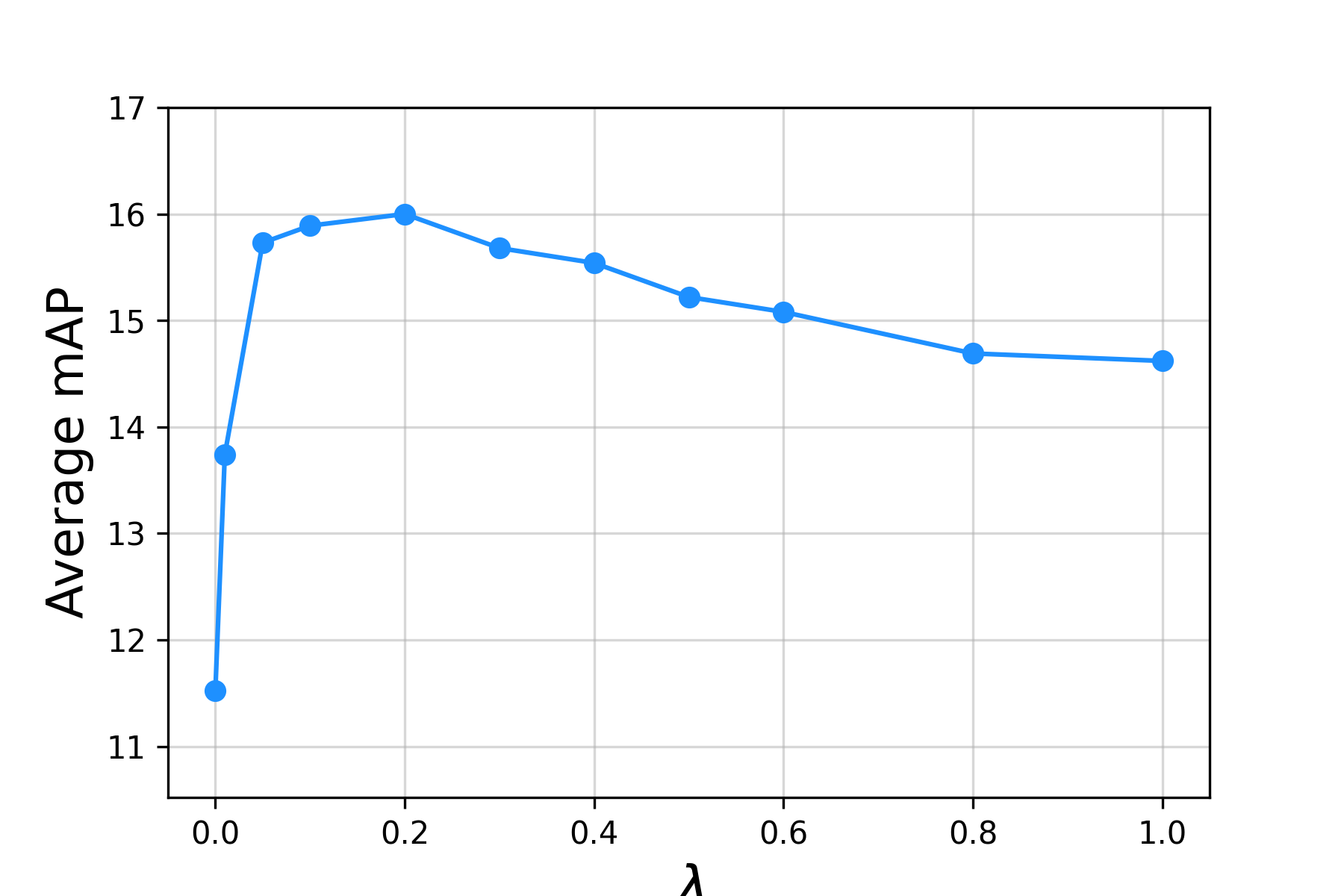}
    \caption{Varying values of the $\lambda$ parameter (a tradeoff between localization confidence and CLIP-based actionness, where SigLIP-SoViT-400M/14 is the CLIP) versus downstream mAP performance on THUMOS14. We hold $\sigma$ to a constant value to observe the tradeoff.}
    \label{fig:lambda}
\end{figure}

\textbf{Effect of actionness scores for suppression.}~ We also ablate the use of actionness scores $\phi^{\text{actionness}}$ via the $\lambda$ tradeoff parameter. $\lambda = 0.0$ corresponds to where only $\phi^{\text{actionness}}$ is used to score intervals, while $\lambda = 1.0$ indicates that $\phi^{\text{actionness}}$ is ignored entirely (i.e. only boundary confidence scores contribute to an interval's score calculation). \Cref{fig:lambda} depicts the effect on average mAP for varying values of the $\lambda$ parameter, keeping all other components of the framework fixed. We observe that $\lambda \approx 0.2$ is the optimal value for the THUMOS14 benchmark, but also that average mAP is greater when $\lambda = 1.0$ than when $\lambda = 0.0$. This implicates two important findings: 1) neither $\phi^{\text{conf}}$ or $\phi^{\text{actionness}}$ alone is sufficient for assessing proposal likelihood (having highly confident boundary predictions and highly confident semantic conent within the interval is not mutually exclusive) and 2) boundary confidence is more critical to performance than CLIP-based actionness within the interval. 

\textbf{Limitations and Future Work.}~The primary limitation of the proposed framework is the computational overhead of inference with \lvlm. This may impact applications where near real-time inference speeds are necessary for utility. However, in many applications, zero-shot localization is a retrospective tool for large-scale video analysis, obviating the need for fast inference. Another limitation of this framework lies in the need to filter relevant action classes to a given sample. When the number of potential action classes is large, this can limit the performance of our framework relying on action-specific queries to \lvlm. One could instead use \lvlm~to generate captions or action classes from key frames of the video itself to limit the size of the query space, and we leave this to future work. 

In addition, the reliance of our method on the logits of \lvlm~requires public access to the model. Also, extracting the logits of the "yes" and "no" tokens as described in this work may not be the only way to assess confidence of \lvlm~in answering visual questions. Instead, one could inflate action classes with key phrases that describe different phases of the action and observe the enrichment of the phrase's token logits at each frame. This strategy and other prompting strategies for \llm~could result in more fine-grained localization outputs, but we leave this as a promising direction for future work.
\section{Conclusions}
\label{sec:conc}

We present \textbf{ZEAL}, a simple framework for true zero-shot action localization in untrimmed video using \llm~and \lvlm. Our method sets a strong baseline for future work in AL where training data is absent or the domain gap between training and inference is large. Furthermore, the examination of \lvlm~outputs (in the form of token logits) as a surrogate for model confidence presents a promising direction for measuring the relevancy of visual inputs to query text, and how this can be used for other tasks like localization. Enabling AL without any training data allows for applications that require fine-grained activity analysis at scale and have prohibitive costs in label acquisition. We hope this work to be the first step towards truly generalizable zero-shot methods for AL on videos in-the-wild.
{
    \small
    \bibliographystyle{ieeenat_fullname}
    \bibliography{main}
}


\end{document}